\documentclass[letterpaper, 10 pt, conference]{ieeeconf}  

\IEEEoverridecommandlockouts                              

\usepackage{graphics} 
\usepackage{epsfig} 
\usepackage{mathptmx} 
\usepackage{times} 
\usepackage{amsmath} 
\usepackage{amssymb}  

\usepackage[colorlinks=true,linkcolor=blue, urlcolor=black, citecolor=cyan]{hyperref}

\usepackage{subfig}
\usepackage{graphicx}
\usepackage{array}

\title{\LARGE \bf MorphoArms: Morphogenetic Teleoperation of Multimanual Robot}

\author{Mikhail Martynov, Zhanibek Darush, Miguel Altamirano Cabrera, Sausar Karaf,   Dzmitry Tsetserukou 
\thanks{
The authors are with the ISR Laboratory, Skolkovo Institute of Science and Technology (Skoltech), 143026 Moscow, Russia.
        {\tt\small \{mikhail.martynov, zhanibek.darush, miguel.altamirano, sausar.karaf,  d.tsetserukou\}@skoltech.ru}}%
}

\begin{document}

\maketitle
\thispagestyle{empty}
\pagestyle{empty}

\begin{abstract}

Nowadays, there are few unmanned aerial vehicles (UAVs) capable of flying, walking and grasping. A drone with all these functionalities can significantly improve its performance in complex tasks such as monitoring and exploring different types of terrain, and rescue operations. This paper presents MorphoArms, a novel system that consists of a morphogenetic chassis and a hand gesture recognition teleoperation system. The mechanics, electronics, control architecture, and walking behavior of the morphogenetic chassis are described. This robot is capable of walking and grasping objects using four robotic limbs. Robotic limbs with four degrees-of-freedom are used as pedipulators when walking and as manipulators when performing actions in the environment. The robot control system is implemented using teleoperation, where commands are given by hand gestures. A motion capture system is used to track the user's hands and to recognize their gestures. The method of controlling the robot was experimentally tested in a study involving 10 users. The evaluation included three questionnaires (NASA TLX, SUS, and UEQ). The results showed that the proposed system was more user-friendly than 56\% of the systems, and it was rated above average in terms of attractiveness, stimulation, and novelty.


\end{abstract}

\section{Introduction}
In recent years, the impact of unmanned aerial vehicles (UAVs) has increased over a wide range of monitoring and inspection tasks. High mobility allows drones to overcome dense obstacles and collect data from hazardous areas. The capabilities of aerial drones are often limited by battery life and the inability to conduct operations on the ground. This does not allow the extensive use of rotor UAVs for long-term and indoor inspections, which often require the ability of the robot to overcome dense obstacles and operate in environments on the ground. To overcome the shortcomings of UAVs, several researchers have proposed combining UAV and unmanned ground vehicle (UGV) designs in flying robots with adaptive morphology, allowing them to switch between motion modes \cite{Zhang_2022}. Bio-inspired design for the problem of motion on water and ground was suggested by Baines et al. \cite{Baines_2022}, achieving locomotion with four morphogenetic legs similar to the turtle limbs. Another bio-inspired design was suggested for the aerial-aquatic hybrid robot developed by Li et al. \cite{Li_2022}, where multirotor UAV was enhanced by a robotic lamellae mechanism.

\begin{figure}[h!]
\centering
 \includegraphics[width=0.9\linewidth]{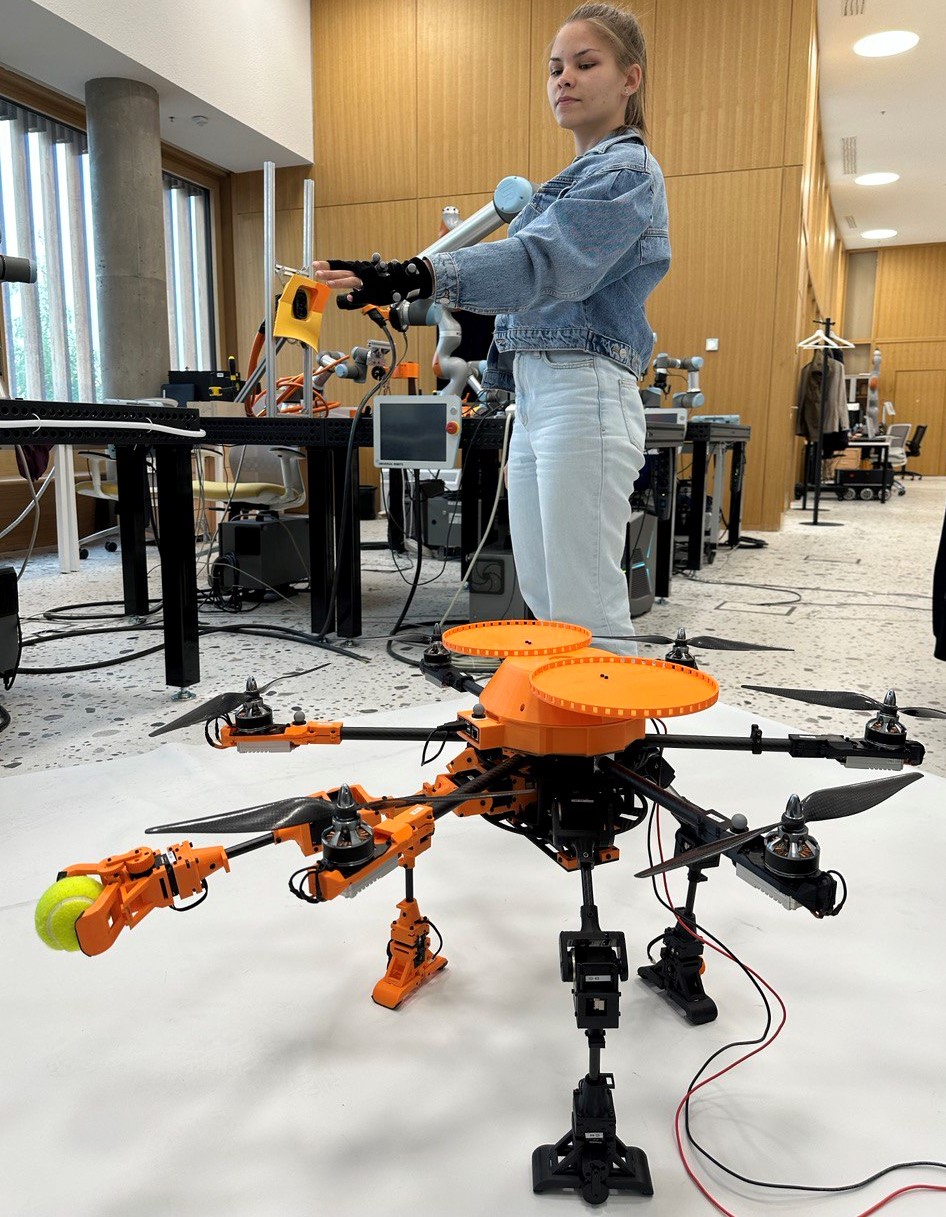}
 \caption{MorphoArms system during telemanipulation.}
 \label{fig:main}
 \vspace{-0.3cm}
\end{figure}

LocoGear \cite{Locogear} research is aimed at investigating the opportunity for locomotion on an adaptive landing platform. The theoretical and experimental analysis of a novel locomotion strategy for a multicopter landing gear with four 2 degrees-of-freedom (DoFs) robotic legs equipped with embedded torque sensors at knee joints were developed. The authors of \cite{Aerovr} proposed a VR-based teleoperation system for a four DoFs robotic manipulator mounted on a UAV. The technology MorphoArms the paper presents, leverages the advantages of both systems, namely that it can land on uneven terrain, and manipulate objects. Additionally, MorphoArms has an advanced kinematic structure of limbs to achieve fast locomotion. The morphogenetic robot is controlled by the operator through gestures.

This work introduces a new morphogenetic UAV chassis and proposes a new concept of walking with the help of four states of pedipulators and using them for manipulation. 
The purpose of this work is to study the controllability of the ground movements and the performance of the manipulation task. The tasks were accomplished through teleoperation using gestures recorded by the motion capture system.
 
In this paper, we present the design and architecture of the platform, as well as the development of a teleoperation control system. To evaluate the performance of MorphoArms, a user study was conducted. Subjects with no previous experience of controlling such a system teleoperated the robot.

\section{Related Works}

A number of research works were focused on the manipulation of objects by drones. For example, Chen et al. \cite{Chen_2022} developed a drone equipped with a robotic arm for aerial manipulation based on multi-objective optimization. A bioinspired perching mechanism with soft grippers was proposed for flapping-wing UAVs by Broers et al. \cite{Broers_2022}. Brunner et al. \cite{Brunner_2022} introduced a planning-and-control framework for aerial manipulation of articulated objects.

Apart from grasping and manipulating objects, robotic pedipulators can work as a locomotion system when integrated with UAVs. Several lightweight legged platforms were previously introduced, e.g., the hexapod robot and four-legged platform ALPHRED, developed by Cizek et al. \cite{Cizek_2018}  and Hooks et al. \cite{Hooks_2018}, respectively.

The work of Zakharkin et al. \cite{Zoomtouch} introduced an innovative approach for robot telemanipulation in the context of real-time control of robots from a teleconference system. The proposed solution employs a DNN for a gesture recognition, which ensures a high level of interactivity between the robot and a human.

The paper~\cite{Visual} proposed a telepresence system for enhancing the aerial manipulation capabilities. It involves not only a haptic device, but also a VR  environment that provides 3D visual feedback for a remotely-located operator in real-time. They achieve this by utilizing onboard visual and inertial sensors, an object tracking algorithm, and a pregenerated object database.


Although the vision-based gesture system has shown a potential as a viable alternative to traditional remote controllers, the recognition of dynamic gestures remains a significant challenge. Yoo et al. \cite{Yoo_2022} proposed a hybrid hand gesture system that combines an inertial measurement unit (IMU)-based motion capturing system with a vision-based gesture system to achieve real-time performance.

\section{System Overview}


The proposed system consists of a morphogenetic UAV chassis and a teleoperation system based on a gesture recognition.  In the previous works, several types of gaits based on inverse kinematics were presented and more details of the robot were introduced in \cite{ICUAS} and \cite{AIM}, respectively. In the present work, a new type of locomotion is presented, so that the robot can rotate on the spot to change orientation for manipulation. In addition, the structure of the grippers, located at the end of each limb, is introduced. 


\subsection{Mechanics}

\begin{figure}[!h]
 \centering%
 \includegraphics[width=1.0\linewidth]{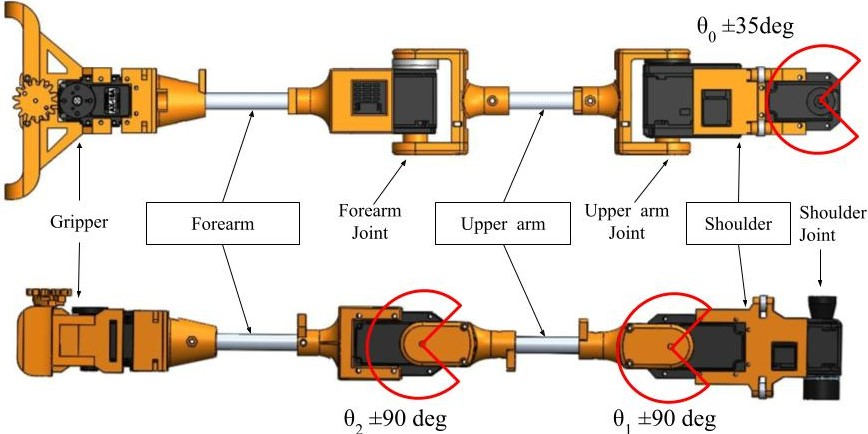}
 \caption{Limb structure of MorphoGear.}
 \vspace{-0.5cm}
 \label{fig:Limb}
\end{figure}

The proposed chassis, dubbed MorphoGear, consists of four multilink limbs with three DoFs. Three different servomotors are used to move the joints of each limb. The upper arm joints of the limbs are driven by Dynamixel MX-106 servomotors (max. torque is of 8.4 Nm), the shoulder joints by Dynamixel MX-28 servomotors (max. torque is of 2.5 Nm), and the forearm joints by Dynamixel MX-64 servomotors (max. torque is of 6 Nm). All parts of the mounts are designed and 3D printed using PLA plastic, and the links are made of carbon. The limb structure is shown in Fig.~\ref{fig:Limb}.

\begin{figure}[!h]
 \centering%
 \includegraphics[width=0.8\linewidth]{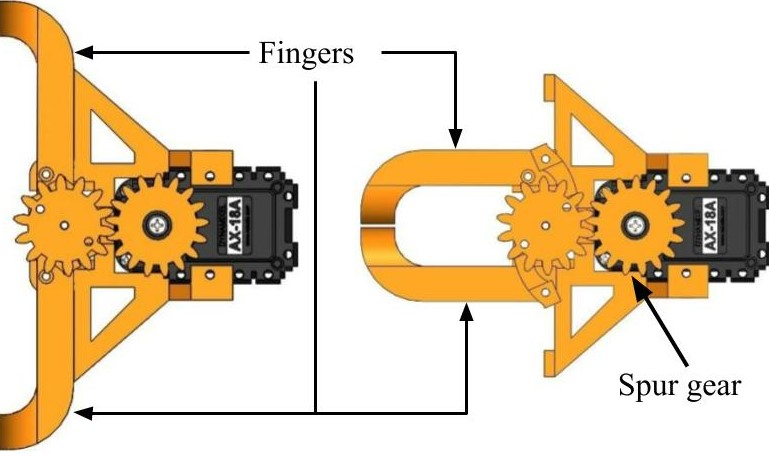}
 \caption{Gripper structure of MorphoGear.}
 \vspace{-0.3cm}
 \label{fig:Gripper}
\end{figure}

Grasping of the grippers is provided by the Dynamixel AX-12A servomotor, which rotates the gear connected by two links to the fingers. Opened and closed gripper is shown in Fig.~\ref{fig:Gripper}

The base consists of two carbon disks, on which the legs and the six axes of the hexacopter are attached. The T-Motor MN4010 rotors are mounted at the end of each axis. The maximum takeoff weight using the six rotors is more than 12 kg, which allows the drone weighing 10.4 kg to fly with payload.

\subsection{Electronics}

MorphoArms system consists of the following parts:
\begin{itemize}
\item Vicon (Mocap System)
\item PC with Unity (Simulation Environment)
\item Raspberry Pi 4 with ROS (Robot Operating System)
\item STM32 (Microcontroller)
\end{itemize}

The motion capture Vicon system tracks the position of the operator's hand, which is transmitted to the game engine Unity, where it turns into a control command for the simulation and the robot. Next, the robot starts executing the command, and the values of the angles of the servomotors are sent to the ROS with the help of the ROS-TCP-Connector Unity package. ROS Node on Raspberry Pi 4 receives commands and joint angles and then sends them through the serial port to STM32, which controls Dynamixel angle of rotation.

\subsection{Motion}
In this paper we consider two modes of motion:
\begin{itemize}
\item Locomotion mode (walking, rotation on place) 
\item Manipulation mode
\end{itemize}

Each limb has the following constraints:
\begin{itemize}
\item Shoulder Joint Angle $\theta_0 \in [-\frac{\pi}{6},\frac{\pi}{6}]$
\item Upper Arm Joint Angle $\theta_1 \in [-\frac{\pi}{2},\frac{\pi}{4}]$
\item Forearm Joint Angle $\theta_2 \in [-\frac{\pi}{2},\frac{\pi}{2}]$

\end{itemize}

The robot design is symmetrical with respect to the center of the base. Thus, it is possible to gait forward, backward, left or right, differing only in the choice of the leading limb. In order to perform a step, the leading limb is selected, the robot locates its two opposite limbs  forward, then lifts itself on them, rotates the shoulder joint, and moves itself forward.

In order to take the most stable position, the forearm must be pointing perpendicular to the ground, so that the following equation must be fulfilled:
\begin{equation}
\theta_1+\theta_2=90^\circ 
\end{equation}

To implement the locomotion, it was chosen to apply a geometrical approach. Since when the shoulder joint is rotated, the limb describes a circle, it is necessary to translate the movement along the circle into movement in a straight line. The equation of the translation in the ground projection is as follows:
\begin{equation}
 \label{shift_horizontal}
 \Delta = 2(1-\cos\tilde\theta(t))l_{UA}\cos \theta_1^{init} ,
\end{equation}
where $\Delta$ is the horizontal shift that occurs while the moving along the circle is translating into a movement in a straight line. Multiplicator 2 means that we compensate two opposite limb rotations, $\tilde\theta(t)$ is the angle to the nearest extreme value of the shoulder joint rotation, it means that for one step the robot puts its limbs forward for $\theta_0=\frac{\pi}{6}, ~\tilde\theta=0$, then lift itself $\theta_0=0, ~\tilde\theta=\frac{\pi}{6}$, then push itself forward $\theta_0=-\frac{\pi}{6}, ~\tilde\theta=0$. $\ l_{UA}\cos \theta_1^{init}$ equals the distance between the shoulder joint attachment point and the end-effector projections on the ground. Here $l_{UA}$ is the length of the upper arm.

The compensation equation on the vertical plane projection:
\begin{equation}
 \label{shift_vertical}
 \Delta = (\cos \theta_1^{init} - \cos (\theta_1^{init}+\xi)) \cdot l_{UA} ,
\end{equation}
where $\theta_1^{init}$ is the initial standing angle, $\xi$ is lifting angle that compensates for shifts in the horizontal plane (ground projection) by decreasing the distance between the shoulder joint attachment point and end-effector projection.

Thus, the system of equations for the locomotion is as follows:
\begin{equation}
 \begin{cases}
 \xi(t) = \arccos((2\cos \tilde\theta(t) - 1) \cdot \cos \theta_1^{init}) -\theta_1^{init}
   \\
\theta_1(t)=\theta_1^{init} + \xi(t)
   \\
\theta_2(t)=\theta_2^{init} - \xi(t)
 \end{cases}
\end{equation}

\begin{figure}[!h]
 \includegraphics[width=1.0\linewidth]{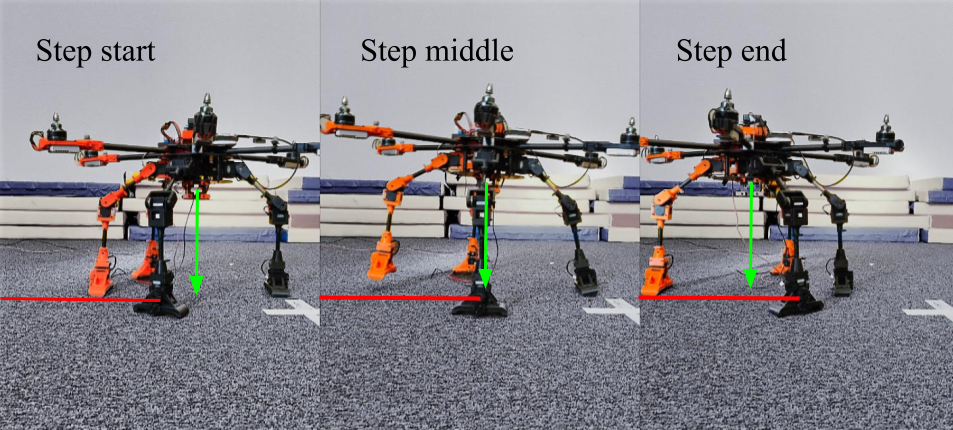}
 \caption{MorphoGear locomotion cycle.}  
 \label{fig:Locomotion}
\end{figure}

The whole locomotion cycle could be seen in Fig.~\ref{fig:Locomotion}. To rotate the robot on place all shoulder joints rotates left or right for $\frac{\pi}{6}$, then the robot steps to initial position one by one limb.

To switch to manipulation mode, the robot puts its opposite limbs forward for $\frac{\pi}{6}$, then its forward limb rises. The MorphoArms manipulation mode stance is shown in Fig.~\ref{fig:main}

\subsection{Teleoperation}

\begin{figure}[!b]
 \centering%
 \includegraphics[width=0.9\linewidth]{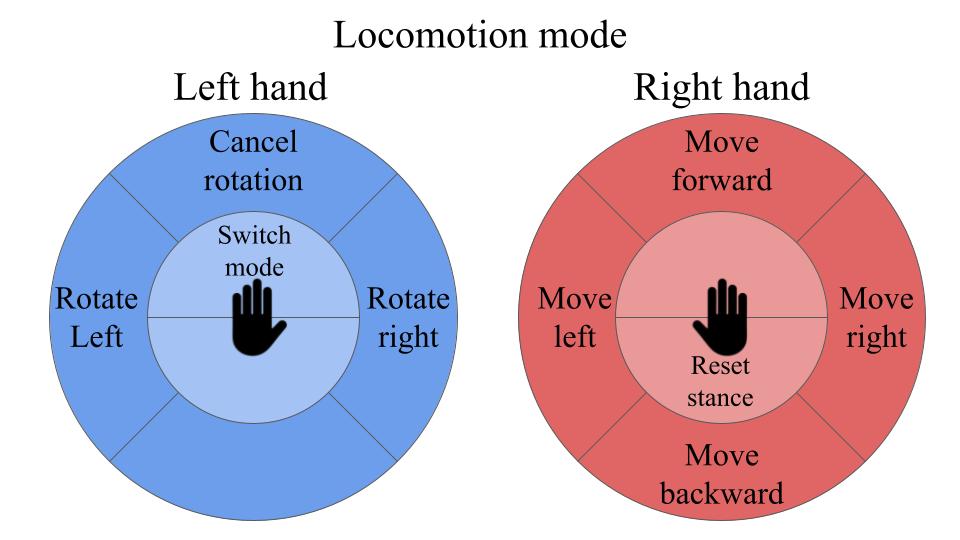}
 \caption{Command bindings for each hand in locomotion mode.}
 \label{fig:MoveCommands}
\end{figure}

\begin{figure}[!b]
 \centering%
 \includegraphics[width=0.9\linewidth]{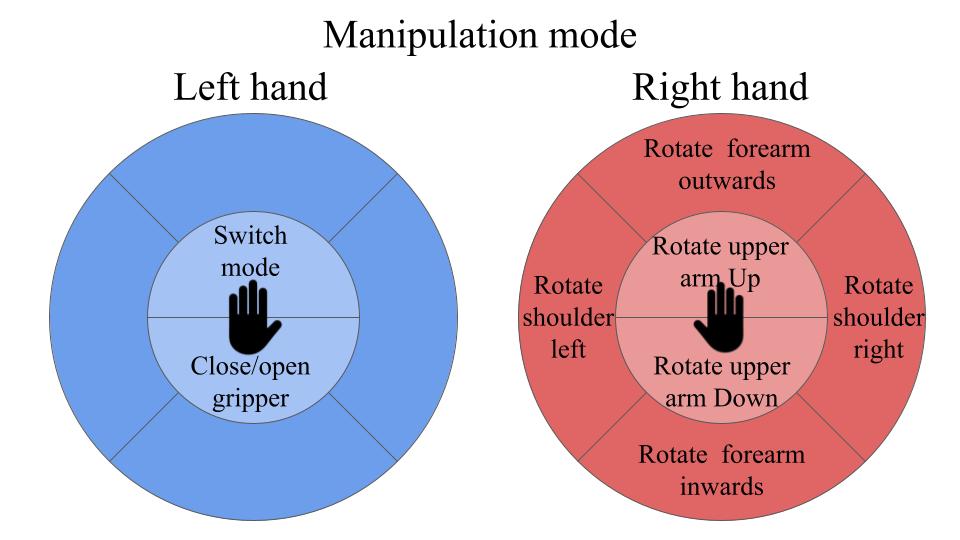}
 \caption{Command bindings for each hand in manipulation mode.}
 \label{fig:ManipulationCommands}
\end{figure}

In this study, we present a system that utilizes a motion capturing approach to enable the control of a robot through the hand gestures. Retroreflective markers are placed on a glove, which is worn by the user, to facilitate the tracking of the hand's position using the Vicon tracking system. The hand's position is subsequently transmitted to the Unity 3D engine through the Vicon Unity package. The hand gestures are translated into specific commands for the robot in Unity. These commands are used as a virtual joystick as shown in Fig.~\ref{fig:MoveCommands} and Fig.~\ref{fig:ManipulationCommands}. Then the required joint angular position is sent directly to the robot through the ROS-TCP-Connector package. By utilizing this system, users are able to intuitively control the robot, allowing for a more natural and immersive human-robot interaction.

Specifically, during locomotion mode, the movement of the right hand is used for moving the robot forward, backward, left and right; and the movement of the left hand is used to rotate in place the robot left and right, cancel the rotation, and switch between locomotion and manipulation modes. During manipulation mode, the right hand controls the position of the manipulator limb, while the left hand is used to open and close the gripper and switch between modes. The control bindings of each hand in each mode are illustrated in Fig.~\ref{fig:MoveCommands} and Fig.~\ref{fig:ManipulationCommands}.

\section{User Study}

A series of experiments were carried out to evaluate user performance while using the proposed system MorphoArms. To evaluate the users' performance, task load, their experience, and the usability of the system, a teleoperation task was designed. During the experiment, users were asked to teleoperate the robot by using gestures.

\subsection{Experiment setup}
\begin{figure}[!b]
 \centering%
 \includegraphics[width=0.9\linewidth]{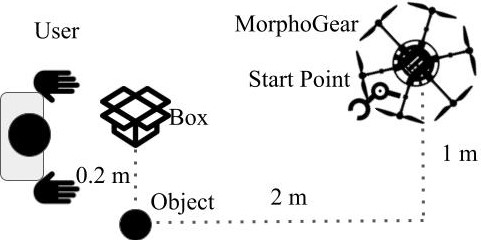}
 \caption{Initial condition of the experiment.}
 \label{fig:Experiment}
\end{figure}

Initially, the robot is placed at a starting point located 2 meters east and 1 meter north of the grasping object. The robot was rotated to the object with its forward limb, and the box was near the object, which was located 0.2 meters north, as shown in Fig.~\ref{fig:Experiment}.
The tasks for the subjects were to:
\begin{itemize}
\item  Walk to the object.
\item  Switch to the manipulation mode.
\item  Grasp the object and locate into the box.
\item  Switch to the locomotion mode.
\item  Get to the initial point.
\end{itemize}

The difficulty was that the robot had to turn to the west so that it could reach the object and put it in the box with one movement. Without turning, the robot would not be able to reach both the object and the box. The experiment is designed in such a way that the participant has to use all the functionality of the robot to complete the task.

\begin{figure}[!t]
 \includegraphics[width=1\linewidth]{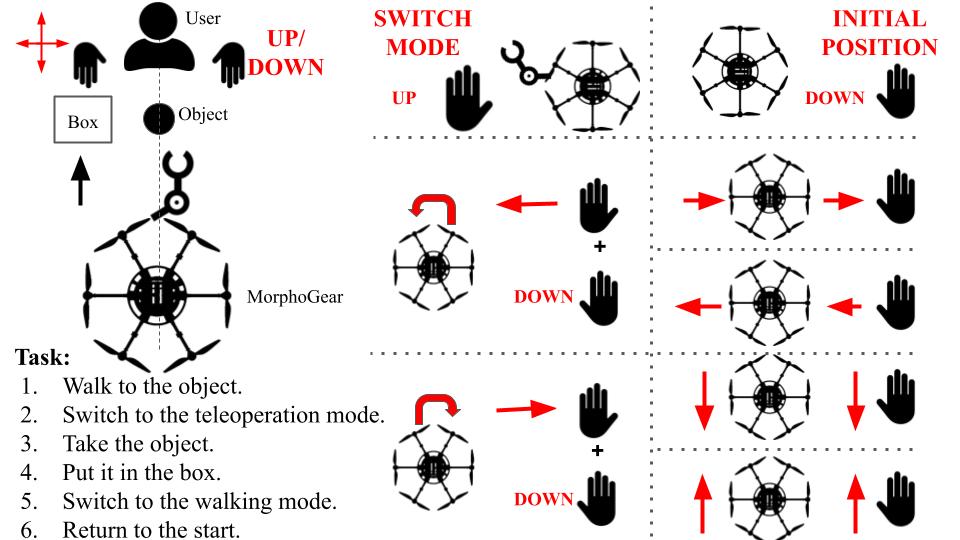}
 \caption{Instructions for the participants of the experiment.}
 \vspace{-0.3cm}
 \label{fig:Instruction}
\end{figure}
\subsection{Participants}
Ten subjects (three females) volunteered to participate in the experiment. The average participant age was 25.1, with a range of 21 to 31 years old. In total, five subjects had never operated robotic arms; three had operated them several times, and the others worked with robotic arms regularly. Five participants had no experience with Vicon, three had used the Mocap system several times, and two reported having regular experience with Mocap.

\subsection{Experimental procedure}
The purpose of the experiment was to navigate the robot along arbitrary trajectories in order to reach the object (a tennis ball) and put it in the box using the proposed system. The robot's movement consisted of rotating the body (left, right, rotation takes 10 seconds), walking in four directions (forward, backward, left, right, each step takes 4 seconds), telemanipulation of one of the robotic arms (3 DoFs, 6 control commands, instant command), grasping (open, close, takes 5 seconds), and switching modes (locomotion, manipulation, takes 15 seconds).
Before the experiment, a training session was performed when the objectives of the experiment were explained, the interface was introduced, and the participants had time to familiarize themselves with the control of the robot using the proposed interface. Graphical instructions were given in order to remind the commands to participants, as shown in Fig.~\ref{fig:Instruction}. 
During the experiment, the users were asked to seat in a chair, to wear the glove with the markers, and to perform the task. If participants failed the manipulation task, they could repeat it until the task was completed. The failure of the task is considered when the ball is thrown out of the robot's reachability area. In this case, the ball returns to its original position, and the participant takes another trial.

\subsection{Measured metrics}

The performance of the experiment was measured by the tasks completion time and the number of attempts. After the experiment, the users were asked to fill out a NASA Task Load Index questionnaire (NASA-TLX) ~\cite{NASA}, a System Usability Scale (SUS) ~\cite{SUS}, and the User Experience Questionnaire (UEQ) ~\cite{UEQ} to analyze each teleoperation experience as a subjective dependent variable. NASA-TLX is a workload rating procedure developed by the Human Performance Group at NASA. It measures the perceived workload of a task over six sub-scales (range 1-20): mental demand, physical demand, temporal demand, own performance, effort, and frustration. The SUS was used to assess the user experience of each interface in terms of three attributes: satisfaction, efficiency, and effectiveness. Ten questions conform to the SUS to rate usability on different scales on a 10-point scale. The UEQ contains 26 questions on a 7-point scale. The UEQ evaluates the experience of the users in six different attributes: attractiveness, perspicuity, efficiency, dependability, stimulation, and novelty. The SUS and the UEQ allow comparison with other types of systems using existing benchmarks ~\cite{Bench_1}, ~\cite{Bench_2}.

\subsection{Experimental results}
During the evaluation, all users were able to complete the task and locate the ball into the box within an average of 2 attempts.
The results showed that the average time to complete the task using the proposed interface was 7:34 minutes. In addition, the average time to walk to an object was 3:14 minutes, and the average time to telemanipulate the object was 2:54 minutes. More detailed data are listed in Table~\ref{tap:speed}.


\begin{table}[]
\centering{
\caption{Experimental results.}
\setlength{\tabcolsep}{2.5pt} 
\renewcommand{\arraystretch}{1}
\begin{tabular}{ | c | c | c | c | c | c |}
\hline
\label{tap:speed}
    \begin{tabular}{c}
    User
  \end{tabular} &
  \begin{tabular}{c}
    Experiment \\
    time, min
  \end{tabular} &
  \begin{tabular}{c}
     Walk \\
     to goal \\
    time, min
  \end{tabular} &
  \begin{tabular}{c}
     Telemani- \\
     pulation \\
    time, min
  \end{tabular} &
  \begin{tabular}{c}
     Trials \\
     number
  \end{tabular} &
  \begin{tabular}{c}
     Walk \\
     to start  \\
    time, min
  \end{tabular} \\ \hline

1 & 7:48 & 3:40 & 2:20 & 2 & 1:48 \\\hline
2 & 7:02 & 4:30 & 1:28 & 1 & 1:04  \\\hline
3 & 5:53 & 1:35 & 2:44 & 1 & 1:34  \\\hline
4 & 7:59 & 3:47 & 2:42 & 2 & 1:30  \\ \hline
5 & 4:15 & 1:58 & 1:10 & 1 & 1:07  \\ \hline
6 & 5:43 & 1:21 & 2:33 & 2 & 1:49  \\ \hline
7 & 6:43 & 3:41 & 1:39 & 1 & 1:23  \\ \hline
8 & 10:28 & 4:05 & 4:22 & 4 & 2:01  \\ \hline
9 & 12:35 & 5:32 & 5:56 & 3 & 1:07  \\ \hline
10 & 7:16 & 2:13 & 4:09 & 3 & 0:54  \\ \hline
\end{tabular}}
\end{table}

\begin{figure}[!h]
 \centering
 {\includegraphics[width=0.80\linewidth]{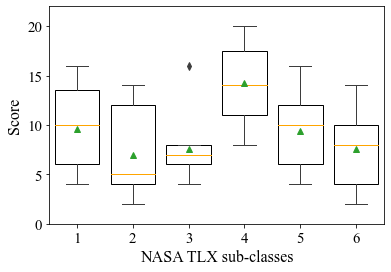}}
 \caption{NASA TLX rating results for the six sub-classes during the teleoperation task, where 1: Mental demand, 2: Physical demand, 3: Temporal demand, 4: Own performance, 5: Effort, 6: Frustration.}
 \vspace{-0.3cm}
 \label{fig:NASATLX}
\end{figure}

The results from the NASA-TLX questionnaire are summarized in Table~\ref{CM1}, and all the results are shown in Fig.~\ref{fig:NASATLX}.

It can be observed that the own performance presents the highest value (14.6) which means that the users considered their own performance at a high level. Results also showed that participants had the lowest physical demand level of 6.9 compared to other criteria.

\begin{table}[h!]
\centering{
\caption{NASA-TLX average rating for the robot telemanipulation.}
\label{CM1}
\setlength{\tabcolsep}{8pt} 

\begin{tabular}{
| p{2.5cm} 
| >{\centering\arraybackslash}p{3cm} |}
\hline
\multicolumn{1}{|c|}{} & \textbf{Results}   \\ \hline
\textbf{Mental Demand}         & 8.6±4.1\\ \hline
\textbf{Physical Demand}            & 6.9±5\\ \hline
\textbf{Temporal Demand}             & 7.4±4.7 \\ \hline
\textbf{Own Performance}          & 14.6±3.4 \\ \hline
\textbf{Effort}            & 9.4±4.1 \\ \hline
\textbf{Frustration}                & 7.1±5.4 \\ \hline
\end{tabular}}
\end{table}

The system Usability Scale reported an average raw score of 70 ± 17.3. According to the percentile rank, introduced by Sauro et al. \cite{Bench_1}, the system is located in a percentile rank of 56\%. Anything with a percentile below 50\% is, by definition, below average, and anything above 50\% is above average. It means that the proposed teleoperation system is above average and more usable than 56\% of the other systems according to the user answers.

\begin{figure}[h!]
\centering
{\includegraphics[width=0.5\textwidth]{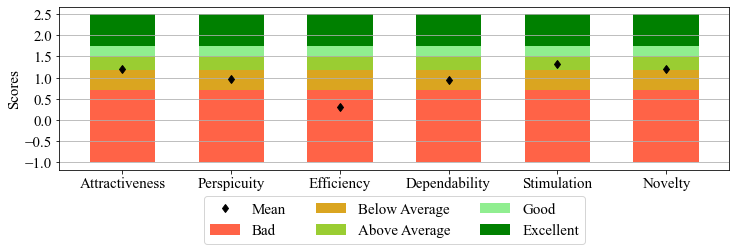}} \\
\caption{UEQ results for the six sub-classes during the use of the proposed system. }
\vspace{-0.3cm}
\label{fig:UEQ_results}
\end{figure}

The results from the UEQ are summarized in Table~\ref{UEQ_table}. Fig.~\ref{fig:UEQ_results} presents the UEQ scores compared with the benchmark values established by Schrepp et al. \cite{Bench_2}. The results of the UEQ revealed, that the proposed system was rated above average in attractiveness (1.2), stimulation (1.33), and novelty (1.2).

\begin{table}[h!]
\centering{
\caption{UEQ rating for the system for the robot telemanipulation.}
\label{UEQ_table}
\setlength{\tabcolsep}{8pt} 

\begin{tabular}{
| p{2.5cm} 
| >{\centering\arraybackslash}p{3cm} |}
\hline
\multicolumn{1}{|c|}{} & \textbf{Results}   \\ \hline
\textbf{Attractiveness}         & 1.2±0.55 \\ \hline
\textbf{Perspicuity}            & 0.98±0.13 \\ \hline
\textbf{Efficiency}             & 0.3±0.65 \\ \hline
\textbf{Dependability}          & 0.95±0.57 \\ \hline
\textbf{Stimulation}            & 1.33±0.58 \\ \hline
\textbf{Novelty}                & 1.2±0.51 \\ \hline
\end{tabular}}
\end{table}

 \subsection{Post-Experience Questionnaire}
After the experiment, the participants were asked about the advantages and disadvantages of using the MorphoArms system and to provide some comments about their experiences. In general, users were very inspired by the MorphoArms system. Some participants noticed that the system is very smooth to control and that it does not allow a new command to be executed until the previous one is finished. The following comments are given as advantages of MorphoArms: ``The control feels natural and intuitive"; ``It feels like I'm just showing the direction to a robot that doesn't know the way"; ``Manipulation with a robot arm looks like a robot is trying to follow your hand movement".
Another comment said that the person was trying to perform the task from the robot's point of view. However, since the participants were sitting opposite the robot, their movements were deliberately mirrored so that when the hand moved to the right, the robot arm also moved to the right from the participant's point of view. This remark will be corrected in future work when the robot's vision system will be developed. Also, in the comments, it was pointed out that in order to read the gesture, the hand must be stretched too far. However, this is also done intentionally to minimize the number of accidental commands. For an experienced operator, this parameter can be reduced.



\section{Conclusions}
A morphogenetic teleoperation system MorphoArms has been developed. It implements a hand gesture recognition system using VICON mocap system. At the local site, the system recognizes hand gestures, which are converted into information to perform the locomotion and manipulation of objects. The performance of the teleoperation task was divided into two stages, walking to the object and manipulation. The average time to complete the task was 7:34 minutes, and the average number of attempts was two. Three questionnaires (NASA TLX, SUS, and UEQ) were used to evaluate the task load, the system usability, and the user experience. Given these results, it can be concluded that the proposed system can potentially be effectively used in a teleoperation task for morphogenetic robots. The results of the NASA TLX questionnaire showed low values in all subclasses, indicating a low mental and temporal demand of the users while using the proposed system. The SUS results rated the usability of the proposed interface above average and identified the possibility of further use of the system for robot teleoperation purposes. UEQ performed well for MorphoArms on such parameters as attractiveness, stimulation, and novelty of the interface. 


In future work, we will improve the teleoperation algorithm and enrich it with new functions such as switching between several types of gait. We will mount an onboard camera for video stream for the operator. In addition, we will use rotors to implement stabilization of the robot during manipulation and walking. 

\addtolength{\textheight}{-12cm}   









\end{document}